# Memory-Augmented LSTM Autoencoder for Unsupervised Activity Recognition with IMU Sensor Fusion

*Saeid Arabzadeh[1], Farshad Almasganj*[2], Mohammad Mahdi Ahmadi[3]*

***Abstract*—**

**Human Activity Recognition (HAR) using Inertial Measurement Unit (IMU) sensors plays a vital role in healthcare monitoring, rehabilitation, and the diagnosis of motor disorders. Despite significant advancements in deep learning for activity classification, major challenges remain such as the reliance on large volumes of labeled data, the complexity of fusing information from multiple sensors, and the limited effectiveness of unsupervised methods in capturing the temporal and spatial dependencies of diverse movement patterns. These challenges are especially pronounced in real-world scenarios where sensor data is noisy, activities overlap, and precise labeling is often unavailable.**

**In this paper, we propose a fully unsupervised spatiotemporal feature fusion framework based on a memory-augmented autoencoder. Our method enhances activity representations using short temporal windows of multi-sensor IMU data, making it suitable for online and real-time applications. The framework first extracts hierarchical static features through a Stacked Autoencoder and fuses them both within and across sensors. These features are then temporally refined using a sequence-to-sequence Long Short-Term Memory Autoencoder (LSTM-AE), which incorporates historical motion patterns without requiring any activity labels.**

**We conduct a comprehensive analysis of key hyperparameters—including sequence length, code size, encoder depth, and memory cell connectivity—and identify configurations that maximize feature separability, even under short-window constraints. We Evaluate our method on two benchmark datasets, DaLiAc and PAMAP2, using realistic inter-class window segmentation. It achieves 96.6% and 98.4% classification accuracy, respectively, surpassing both supervised baselines and existing unsupervised approaches. Notably, our method improves feature separability by up to 9%, despite operating with significantly shorter temporal windows. It is worth noting that our experiments show inter-class window segmentation can reduce classification accuracy by approximately 7%. However, this approach was intentionally adopted in our study to better reflect real-world activity transitions and enhance the model's practical relevance.**

## 1. INTRODUCTION

HUMAN Activity Recognition (HAR) has attracted significant attention in recent years due to its wide range of applications, including fall detection, security systems, wearable devices, rehabilitation, and the assessment of neuromuscular diseases such as Parkinson's and Alzheimer's [1-4]. Moreover, with the aging population and the rising prevalence of chronic illnesses, the demand for smart healthcare systems particularly those based on wearable sensors has become more critical than ever [5 ,2]. Among the different sensing technologies, wearable inertial sensors such as accelerometers and gyroscopes have attracted the most attention in HAR research, largely due to their low cost and ease of deployment, and this study focuses specifically on this approach [6].

Despite the broad applications and extensive research in this domain, HAR still faces several challenges. These include the difficulty of labeling data, scarcity of data across diverse activities, and the high similarity between different movement patterns [8 ,7]. Furthermore, HAR performance can be severely affected by movement disorders caused by neurological and non-neurological factors (e.g., aging, weight gain), and by practical issues like sensor displacement [9]. To address these issues, numerous Deep Neural Network (DNN) based HAR models have been proposed, primarily aiming to extract informative features from multi-sensor inputs and fuse them effectively to recognize activity patterns. These models generally perform well under controlled settings with clean datasets and minimal signal disruption [10 ,4].

Among DNN architectures applied to HAR, convolutional neural networks (CNNs) have been particularly prominent, due to their spatial feature extraction capabilities within fixed time windows [11-13 ,4]. For instance, Kobayashi et al. proposed MarNASNets, a deep CNN with over 15 layers optimized for sensor-based feature extraction and classification [4]. Similarly, Yang et al. developed a multichannel CNN framework to extract both spatial and temporal features and enhance performance through manual feature engineering [12]. Although CNN-based models laid the foundation for deep learning in multichannel HAR, they require large amounts of data to train each class and struggle to differentiate similar activities within a short time window [12]. Additionally, Their performance declines when distinguishing a wide range of complex movements, especially in clinical or rehabilitative contexts, where accurate interpretation is critical [15 ,14]. To mitigate these limitations, some studies have attempted to extend input window length, allowing the model to capture long-range temporal features [16]. However, experimental results indicate that increasing window length alone reduces model practicality and is insufficient without architectures that explicitly model the temporal dependencies. Consequently, recurrent neural networks (RNNs), especially Long Short Term Memory (LSTM)

1, 2 and 3 is with the Faculty of Biomedical Engineering, Amirkabir University of Technology (Tehran Polytechnic), Tehran, Iran
e-mail: 1. arabzadehsaeed@aut.ac.ir, 2. almas@aut.ac.ir , 3.mmahmadi@aut.ac.ir)

and Gated Recurrent Unit (GRU) variants, have gained attention in HAR [17 ,16]. For example, Rojanavasu et al. highlighted CNNs' limitations in modeling temporal dependencies and evaluated various RNN lengths, concluding that hybrid CNN and LSTM architectures outperform standalone of these models [17]. These models treat windowed signals as time sequences and extract features across these sequences. While they improve recognition accuracy over time, their large number of parameters poses challenges for supervised training, especially when dealing with many activity classes. Moreover, these networks focus more on temporal transitions between windows than momentary features, which limits their performance in capturing transient cues [18]. Since 2022, there has been increasing interest in hybrid CNN–RNN models that capture both spatial and temporal patterns. These have been implemented in sequential [19-23 ,15 ,11], parallel [24], and multi-stream [25] configurations. For instance, Koşar and Barshan utilized a dual-branch model combining raw data and spectrogram inputs processed through 1D and 2D CNNs, followed by LSTM layers [13]. Similarly, Baloch et al. compare early vs late-fusion CNN–LSTM architectures model. In this research, they investigated different feature extraction structures from each sensor, multiple sensors, and all sensors using CNNs, and they fuse extracted features using CNN-LSTM networks. They showed, late-fusion CNN–LSTM model improved subtle activity recognition over early-fusion designs [14]. In another study, two LSTM layers classified Parkinsonian movements into four severity levels by CNN-based spatial features. In this model, the spatial and temporal features were fused in the Dense Layers Classification (DLC) stage [24]. More recently, Gaud et al. proposed MHCNLS-HAR, a hybrid model combining multi-headed CNNs, LSTMs, and Attention Mechanisms (AM) for real-time gesture recognition on edge devices [25]. These hybrid models aim to extract complex intra-frame features as dynamic proxies and merge multiple sequential patterns using RNNs. Although these studies improved dynamic feature modeling, they rely heavily on supervised classification losses and require long sequences during training. Their performance also depends on careful layer tuning and accurate labeling, which hinders generalization on real-world noisy data [19-22 ,15 ,11]. Furthermore, they often struggle to fuse sensor data from varied body positions and are sensitive to the number and placement of sensors [26].

To overcome sensor-level fusion challenges, several studies have adopted AM and transformer-based architectures. Given CNNs' prevalence in HAR, Squeeze-and-Excitation (SE) is one of the most common attention modules in this field. It calibrates channel-wise feature importance and allows the network to selectively emphasize more relevant sensor data [28 ,27]. For example, Wang et al. extracted features from an axis signal and all sensors signals by multi branch CNN and applied SE-based attention across CNN branches [29]. Similarly, Zhang et al. integrated SE blocks into a CNN–LSTM model for sequential feature refinement [27]. In a more advanced design, Essa, and Abdelmaksoud, introduced a temporal-channel CNN with self-attention, inspired by transformers, to capture both temporal and inter-channel dependencies [30]. The goal in attention-based structures is to assign value to features, whereas in processing multiple IMU sensors, the primary objective is featuring fusion and processing the co-activity between them, not the analysis of data for an overall representation. Generally, previous studies indicate that in scenarios where multiple IMU sensors are used for a variety of human movements, the performance of attention mechanism-based structures tends to decrease compared to hybrid CNN and LSTM structures [19].

Unsupervised autoencoders (AEs) have been proposed to address challenges like data scarcity, sensor fusion, and class imbalance. However, their extracted features often suffer from limited discriminability, especially, when dealing with a wide range of complex human activities [15]. Many proposed AE-based HAR models, directly combine CNN and LSTM layers without adequately fusing spatial and temporal features or accommodating complex sensor interactions, resulting in reduced accuracy when activities or sensor numbers increases.

In this paper, we propose a novel method based on LSTM architecture and autoencoders, which enhances the static features extracted from human activity (HA) within a single framework by incorporating information from past frameworks. We refer to this method as the **Unsupervised Temporal Feature Fusion (UTFF)** approach. It consists of two main components: (1) extraction of static features, and (2) updating those features using previously observed data, when available.

In the first part, we introduced a hierarchical model named **Hierarchical Unsupervised Fusion (HUF)** [31], which performs feature extraction and fusion across three levels. Initially, the information from each framework is projected into a high-dimensional feature space by designing a stacked CNN-based autoencoder. In the second and third stages, features are fused intra-sensor and subsequently across all body-worn sensors using CNN-AE structures. The second part, which constitutes the core contribution of this work, performs temporal feature fusion using a memory-based autoencoder built on LSTM networks. This component combines features from the current framework with those from previous time steps. Through this architecture, we address several key challenges: the scarcity of labeled data required for training deep supervised models, the need for effective sensor fusion in multi-sensor setups, and the ambiguity of motion start and end boundaries within a given time window. Our model leverages dynamic feature extraction in an unsupervised sequence-to-sequence (STS) learning framework to adaptively integrate current and past information.

The UTFF method is versatile and applicable to a wide range of HAR settings from single device smartphone-based systems to full-body multi-sensor configurations. Moreover, applying unsupervised learning to HAR datasets with a large number of activity classes is a significant challenge. To our knowledge, this is the first study that simultaneously addresses both sensor-level and temporal fusion in an unsupervised manner, utilizing a hierarchical structure with independently trained modules. Owing to its flexibility across datasets with varying sensor configurations, the proposed model holds strong potential for practical real-world applications.

Our main contributions are:

1- We introduce the UTFF framework, which performs both static and dynamic feature extraction in an unsupervised manner.

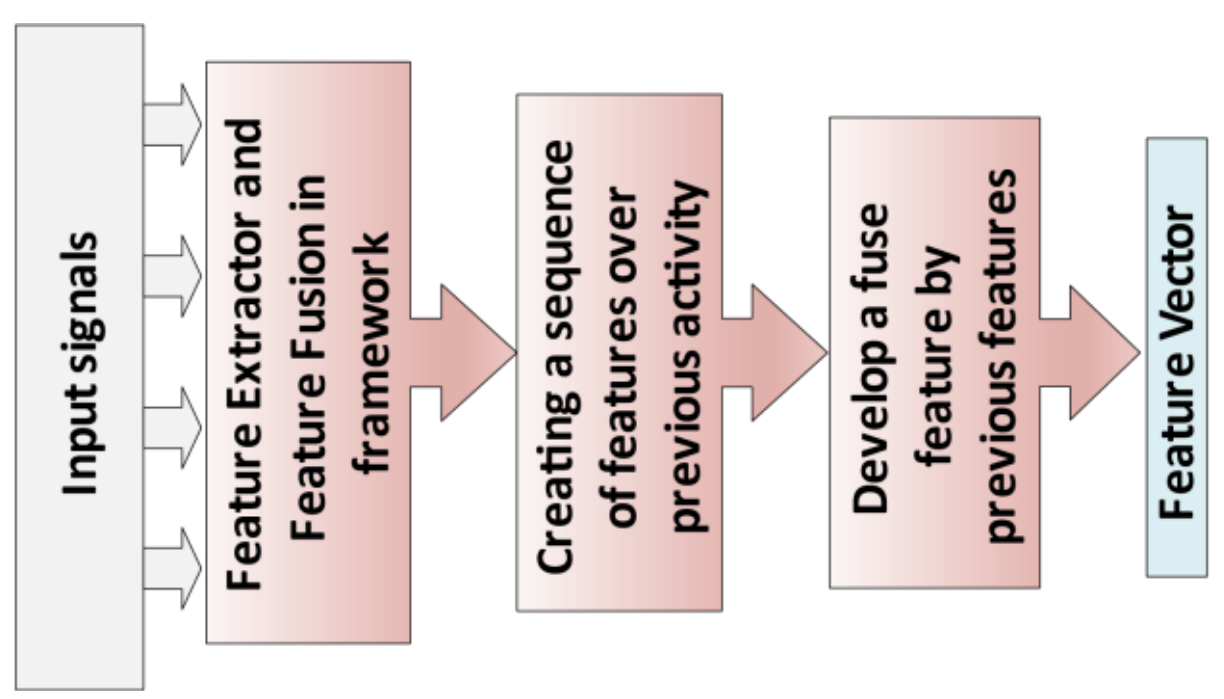

*Figure 1 The unsupervised processing stages of the UFF-ST method, from raw signal acquisition to final feature extraction for human activity classification*

Owing to its hierarchical structure, the model extracts and fuses features from the sensor level up to the full-body level. More importantly, the static features are updated using historical features without any supervision. This approach addresses several key challenges, including data scarcity, signal misalignment and labeling errors, multi-sensor fusion, unsupervised dynamic feature extraction without requiring long-term sequences, and generalization across heterogeneous sensors.

2- We present a memory-enhanced autoencoder that updates current framework features by incorporating relevant information from preceding frameworks. This architecture not only preserves the current context but also adapts based on historical data when available. It specifically targets the problem of limited data availability in HAR for training models with a large number of activity classes in unsupervised manner by RNNs network.

3- Unlike previous works, our model explicitly handles uncertainty in activity transitions during windowing. By designing for short-duration dynamic feature extraction and incorporating complete training for such cases, the model minimizes errors associated with imprecise window boundaries. This design choice aligns more closely with real-world HAR conditions.

4- We systematically assess how variations in window size, feature dimensionality, and temporal sequence length affect the separability of learned dynamic features. The effectiveness of these features is further evaluated using a linear classifier.

5- In contrast to conventional intra-class windowing that segments data within individual activity classes, our method segments frameworks continuously across the entire signal timeline. This strategy allows multiple activities to appear within a single window. We provide a comparative analysis of feature extraction results, both static and dynamic, under this more realistic windowing scheme.

## 2. Methods

In this study, we propose a model based on autoencoders and the concept of STS learning to address the need for unsupervised spatial and temporal feature extraction and fusion from multiple IMU sensors in HAR applications [32]. As we show in Figure 1, initially, the model extracts feature from a short-term framework, then performs intra-sensor and inter-sensor fusion across the body-worn sensors. The resulting fused feature vector is treated as a sequence of movements, which is then processed by a memory-based autoencoder to reconstruct each current framework's features using information from previous frameworks. Finally, to evaluate the quality of the extracted features, we employ a DLC for classification. In the following sections, each stage of the proposed approach is described in detail.

### 2.1. Feature Extraction and Feature Fusion in farmwork

*Inputs of networks*

IMU sensors typically provide six motion channels from the accelerometer and gyroscope, which for an activity segment of $N$ samples, the data from each sensor can be represented as:

$[a_x[n], a_y[n], a_z[n], g_x[n], g_y[n], g_z[n]] \quad$ where $n \in [0, N]$

If there are $s$ IMU sensors deployed, the combined sensor data can be structured as a tensor $X \in R^{s \times 6 \times N}$, organized as:

$$X = \begin{bmatrix} a_{x_1}[n] & a_{y_1}[n] & a_{z_1}[n] & g_{x_1}[n] & g_{y_1}[n] & g_{z_1}[n] \\ \vdots & \vdots & \vdots & \vdots & \vdots & \vdots \\ a_{x_s}[n] & a_{y_s}[n] & a_{z_s}[n] & g_{x_s}[n] & g_{y_s}[n] & g_{z_s}[n] \end{bmatrix}$$

Each element of the matrix $X$ is denoted by $x_i[n]$, and feature extraction and fusion are performed in three steps to interpret the recorded motion pattern (Figure 2-a). These steps are illustrated in Figure 2, and we briefly outline each of them in the following subsections. For a more detailed analysis, we refer the reader to our previous work in [31], in which we introduced a HUF method. This method performs hierarchical feature extraction and fusion from individual sensor channels up to all sensors placed on the body.

As shown in Figure 2-b, each signal axis is fed into the Data Representation Stacked Autoencoder (DR-SAE), whose output is then fused intra-sensor using the Local Feature Fusion Autoencoder (LFF-AE) network. Finally, the Global Feature Fusion Autoencoder (GFF-AE) network performs inter-sensor fusion on the outputs of the LFF-AE to produce a single, unified feature vector that represents the static features of the activity over the NNN input samples.

- Step one - DR-SAE:

As illustrated in Figure 2, each $x_i[n]$ is processed through the DR-SAE Network that embedded to a higher-dimensional feature space. This network consists of five convolutional layers, each employing the SELU activation function. It disentangles the information embedded in the raw signal and represents it as 256 new signal channels. The autoencoder is designed in an overcomplete configuration and trained using a six-stage stacked autoencoder approach as shows in Figure 2 by read arrows in DR-SAE network. The feature vectors extracted from the output layer (OL), encode distinct patterns of the motion signals recorded in each channel. This architecture enables unsupervised fusion of features both within and across sensor channels, significantly improving the outcomes of the next two steps in the pipeline. The detailed structure of the DR-SAE can be seen in Figure 2-c, where the number of input and output channels for each layer is shown.

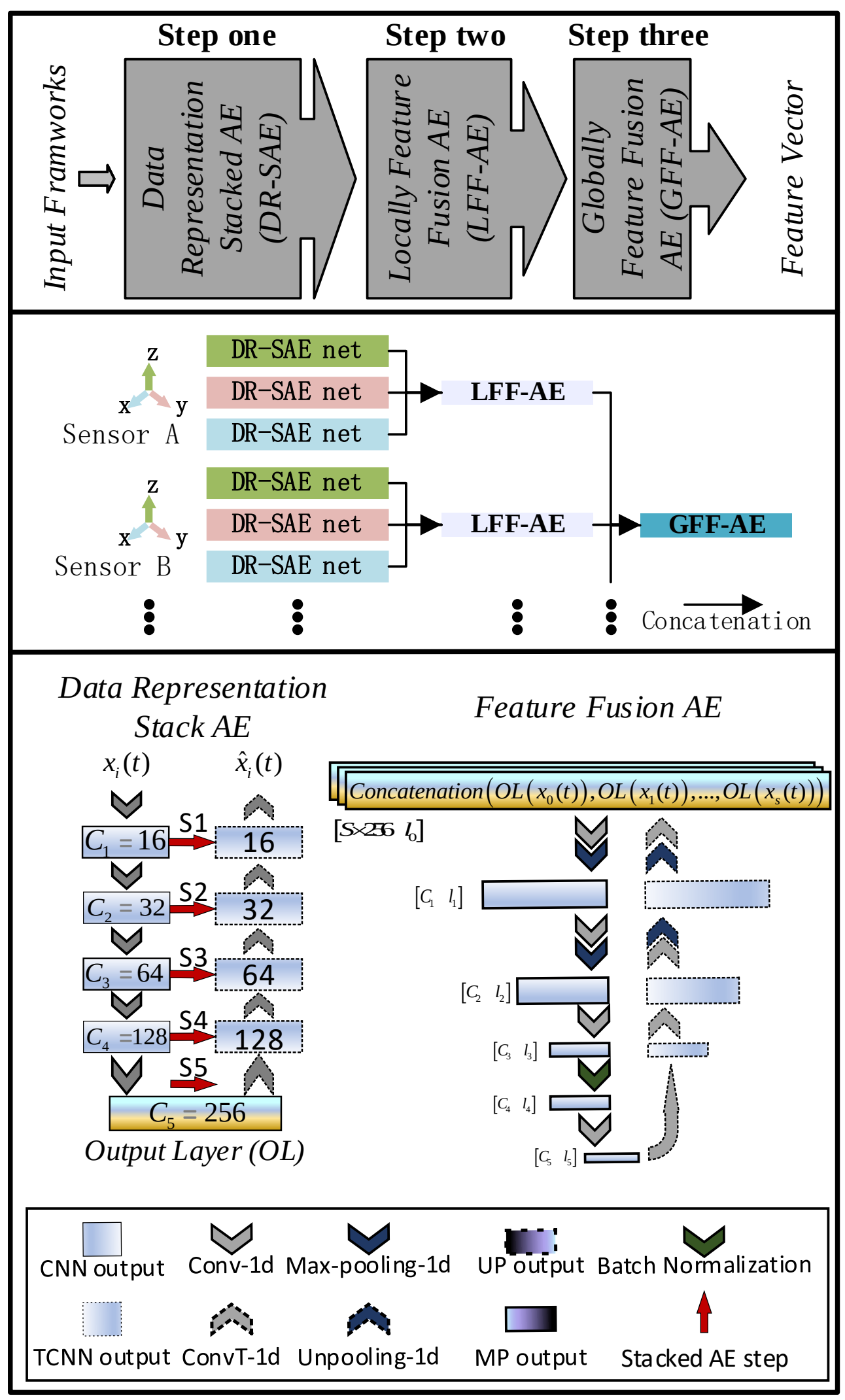

*Figure 2 The HUF model performs static feature extraction from a single time window in three hierarchical steps. The term* Feature Fusion AE *refers to the general architecture shared by both the* Local Feature Fusion Autoencoder (LFF-AE) *and the* Global Feature Fusion Autoencoder (GFF-AE) *networks.*

- Step two – LFF-AE:

LFF-AE is composed of eight convolutional layers, two max-pooling layers, and one batch normalization (BN) layer, and it operates in the form of an autoencoder as illustrated in Figure 2-c. The main purpose of this step is to fuse intra-sensor features extracted by the DR-SAE networks (denoted as OL in Figure 2-c). An example of the DR-SAE output for a sensor consisting of both an accelerometer and a gyroscope can be organized into a matrix and concatenated as follows, which is used as the input to the LFF-AE:

$$LFF-AE_{Input} = Concatenation(OL(a_x[n]), OL(a_y[n]), OL(a_z[n]), OL(g_x[n]), OL(g_y[n]), OL(g_z[n])) \quad \text{Equation 1}$$

According to Figure 2-c, a max-pooling layer follows each of the first two convolutional layers in the LFF-AE. These pooling layers help stabilize the model against motion disturbances, such as sudden acceleration due to hand-body contact during walking or variations caused by sensor displacement throughout movement [33]. Additionally, a batch normalization layer is applied after the third layer, which not only helps regulate signal magnitude but also accelerates the training of the autoencoder. The encoder and decoder architectures in this network are symmetric, with the exception that batch normalization is not applied in the decoder. Finally, the hyperparameters of this network were selected based on their impact on the final feature vector from GFF-AE and the classification performance for the static features, as summarized in Table 1.

- Step three – GFF-AE

In the third step of the HUF method, we perform inter-sensor feature fusion. At this stage, a unified feature vector is generated that represents the subject's movement based on data collected from all body-worn sensors. Due to the nature of this fusion process, we refer to this step as the Global Feature Fusion Autoencoder (GFF-AE). Although the GFF-AE shares a similar architecture and layer configuration with the LFF-AE, its model parameters differ. These differences optimized for global feature integration are detailed in Table 1.

*Table 1 LFF-AE and GFF-AE channels number from input to the output of encoders. Letter A shows the number of axes in one sensor unit and S shows the number of sensors.*

| ***Fusion network*** | $C_0$ | $C_1$ | $C_2$ | $C_3$ | $C_4$ | $C_5$ |
|---|---|---|---|---|---|---|
| **LFF-AE** | A*256 | 256 | 256 | 256 | 256 | 256 |
| **GFF-AE** | S*256 | 256 | 128 | 128 | 64 | 64 |

### 2.2. Making Frame work to Sequence

As previously described, each input framework contains $N$ samples with a 50% overlap between successive windows, the sequences can be denoted as $w_1, w_2, w_3, \ldots$ as illustrated in Figure 4. After segmentation and tensor formatting of all sensor inputs, each window is processed using the HUF method to extract features. The resulting feature vectors, denoted by $F_{w_1}, F_{w_2}, F_{w_3}, \ldots$ are shown in Figure 4. These vectors are then concatenated to form a temporal sequence of motion features as follows:

$$Features\ Vectors = [F_{w_1}, F_{w_2}, \ldots, F_{w_k}, \ldots] \qquad \text{Equation 2}$$

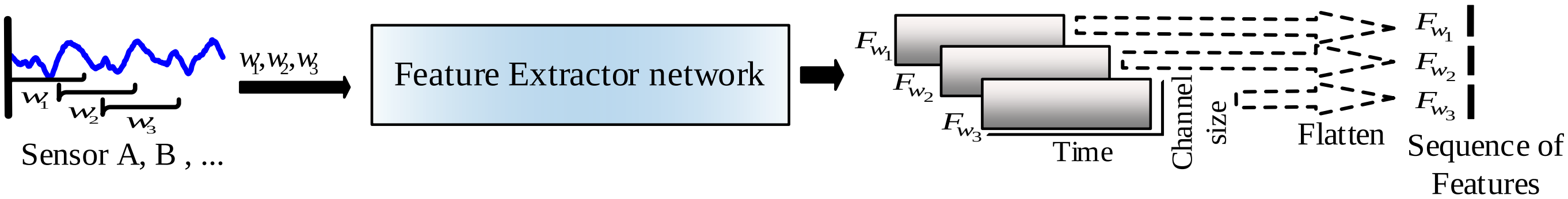

*Figure 4 Forming a sequence from static features of HUF method- Each input window ($w_i$) contains multi-channel signals from different sensors. After being processed by the feature extraction network, these inputs are transformed into a feature vector denoted by $F_{w_i}$.*

where $k$ is the number of extracted segmented frame work that are recorded for a given subject.

In this study, we introduce a memory-based autoencoder built on the STS structure. This model develops each feature vector with its preceding $q$ vectors and enhances the current feature using historical context. To implement this, the first sequence is initialized as $[0,0,\ldots,F_{w_1}]$ with a shape of $f \times q$ where $f$ is the number of features. In the first sequence, $q-1$ vectors are zero due to we do not have information for them and for more reality we consider all of these situation in our train and test. Similarly, the second sequence is formed as $[0, 0, \ldots, F_{w_1}, F_{w_2}]_{f\times q}$ where only the last two columns contain actual feature vectors and this pattern continues for each subsequent input.

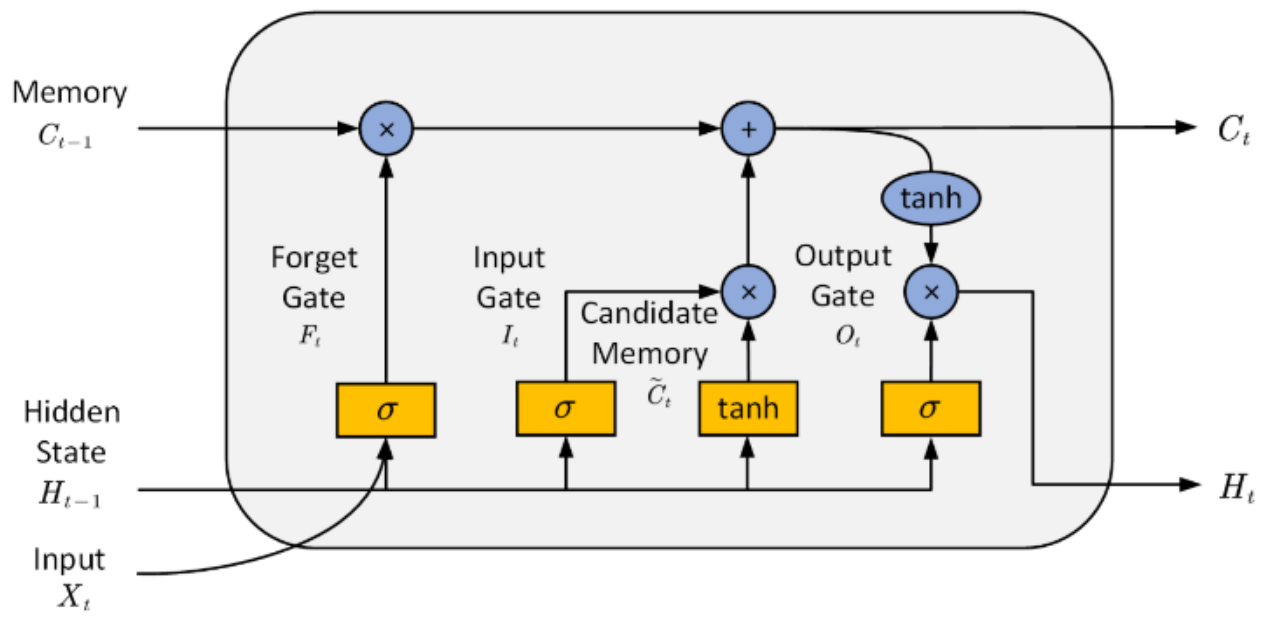

*Figure 3 General Structure of an LSTM Cell – $X_i$: input at the i-th step of the sequence, $H_{i-1}$: output of the previous cell, $C_{i-1}$: cell state from the previous step, $\sigma$: sigmoid activation function, $+$: element-wise addition, $\times$: element-wise multiplication (Hadamard multiplication).*

## 2.3. Memorized Auto Encoder by Seq-to-Seq modeling

An LSTM network is a specific type of RNN with a specialized architecture that enables it to retain both short-term and long-term memory. Leveraging this memory capability and its strength in carrying information across sequences, we design an architecture to extract dynamic features from the HUF model's output. In general, LSTM networks are commonly used for dynamic data classification [34]. However, we show in an autoencoder structure, it can capture dynamic representations that enable reconstruction of output sequences. Although LSTM-based models have not previously been used for unsupervised feature fusion, we propose such a method for HAR, in which static features from each framework are enhanced using past time steps via an LSTM-based memory structure.

The LSTM unit consists of four gates, as illustrated in Figure 3. These gates are implemented using feedforward networks, and by varying the activation functions and input types, the architecture is able to maintain memory over time. At each time step, the LSTM cell receives three matrices: $X_i$ (current input feature), $H_{i-1}$ (output of the previous cell), and $C_{i-1}$ (previous cell state).

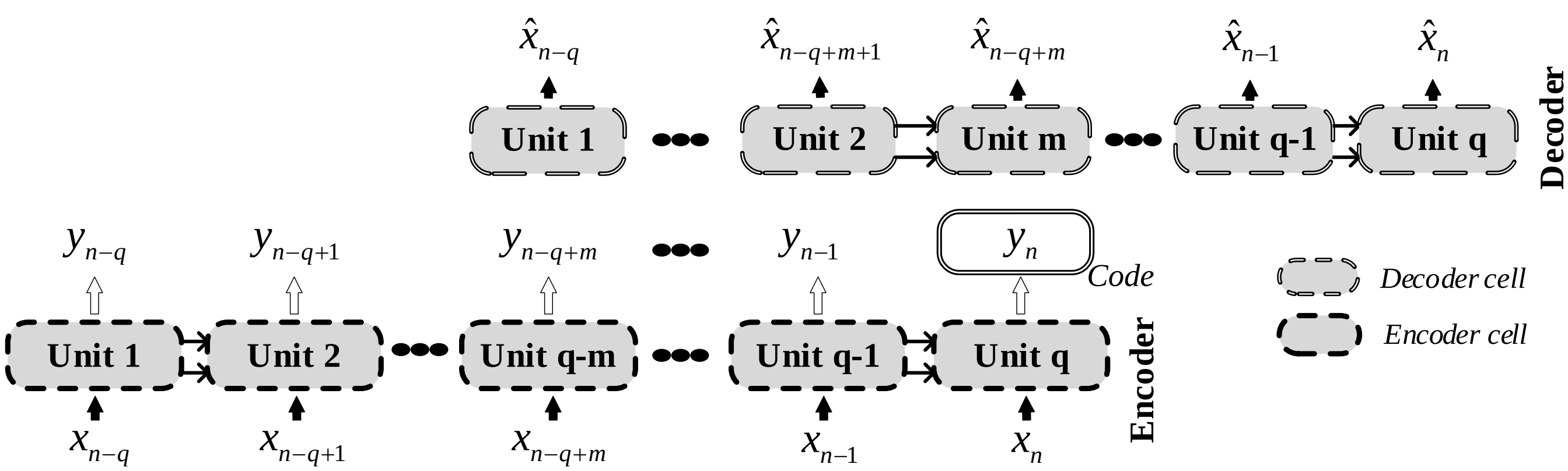

*Figure 5 Proposed Memory-Based Autoencoder Model – $x_n$: feature vector of the n-th framework, q: number of frameworks in the input sequence from previous time, m: number of frameworks used for signal reconstruction in the decoder, $\hat{x}$: reconstructed feature vector.*

The outputs are $C_i$ (current cell state) and $H_i$ (current hidden state), which are forwarded to the next time step. The main equations for the LSTM cell, as shown in Figure 3, are:

$$F_t = \sigma(W_f * concat(h_{t-1}, x_t) + bf) \quad \text{Equation 3}$$

$$I_t = \sigma(W_I * concat(h_{t-1}, x_t) + b_I) \quad \text{Equation 4}$$

$$C_t = tanh(W_c * concat(h_{t-1}, x_t) + b_c) \quad \text{Equation 5}$$

$$O_t = \sigma(W_o * concat(h_{t-1}, x_t) + b_o) \quad \text{Equation 6}$$

$$CS_t = (CS_{t-1} \times F_t) + (I_t \times C_t) \quad \text{Equation 7}$$

$$h_t = O_t \times tanh\,[(CS_{t-1} \times F_t) + (I_t \times C_t)] \quad \text{Equation 8}$$

where $F_t$ is forget gate, $I_t$ is input gate, $C_t$ is candidate, $o_t$ is output gate, $CS_t$ is cell state, and $h_t$ is hidden state or final output. Also, $*$ denotes matrix multiplication, $\times$ represents the Hadamard product, $b$ is the bias vector, and $\sigma$ is the sigmoid activation function.

In this study, by connecting multiple LSTM cells as shown in Figure 5, we constructing a memory-based autoencoder capable of extracting features from sequences of length $q$. The model works AE structures which is updates the final cell's features based on previous cells such that the input sequence is reconstructed solely from the information retained in these final states. In the encoder and decoder, $q$ LSTM cells are used. Of these, the final $m$ cells of the encoder are connected to the first $m$ cells of the decoder. The model is trained in an **STS** manner, aiming to reconstruct the original input sequence of length $q$ from a compressed representation. To achieve this, the encoder must condense the sequence's key information, especially from the $q - th$ element, into a compact representation for the decoder to reconstruct. This mechanism allows temporal information to be aggregated over time, and the final cell feature is enhanced based on its preceding steps.

The decoder's input consists of the final $m$ outputs of the encoder, while the remaining positions are initialized with zero matrices. As discussed in Section III.D, choosing a smaller value for $m$ encourages the encoder to generate richer features, as the entire sequence must be reconstructed from fewer intermediate representations. We also explore the effects of changing the values of $q$, the output feature dimension, and the number of layers in the encoder. All initial memory states in both the encoder and decoder are set to zero.

The LSTM-AE training procedure is summarized in Algorithm 1. The model's parameters Θ include all weights and biases of the encoder and decoder LSTM cells. The objective is to minimize the reconstruction loss between the original input

and the reconstructed input, using the mean square error loss function:

$$loss = \frac{1}{q}\sum_{i=1}^{q}(x_i - \hat{y}_i)^2 \quad \text{Equation 9}$$

In this paper, we refer to the final hidden state ($h_q$) as the code, which represents the enhanced dynamic feature vector**.** As illustrated in Figure 6, each neuron in $h_q$ consists of contributions from both the current input and the output of the previous cell. After applying the sigmoid activation and weighting based on the cell's gates and prior state, the resulting value is scaled using a weight matrix $k_f$. By analyzing the interaction between the vectors $F_{x_q}$ and $h_{q-1}$ by linear layer, we can quantify how information from previous time steps contributes to the final representation of framework $q$. Each neuron in $h_q$ integrates past information with all current features through weighted summation, forming a new enhanced feature representation. Given this mechanism, the dimensionality of $h_q$ plays a critical role in both computational complexity and the effectiveness of information transfer from past features to the current representation. Therefore, in **Section –**, we analyze the impact of this key parameter on the classification performance of individual activity classes as well as on the overall quality of the extracted features.

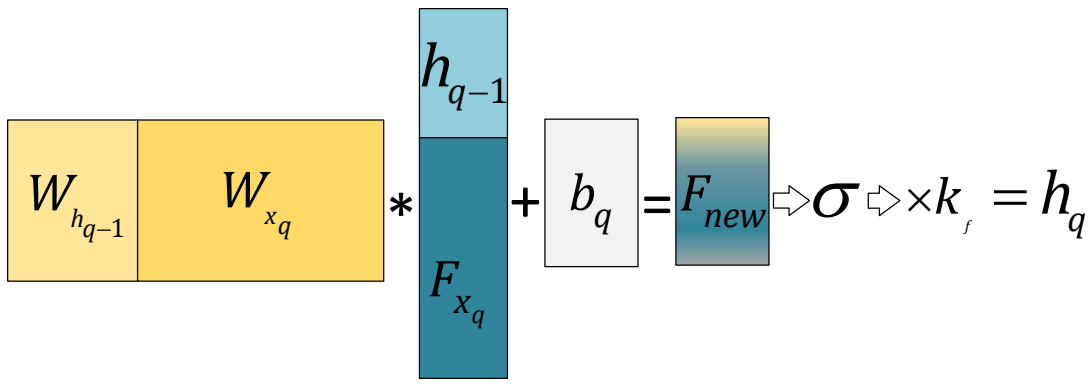

*Figure 6 illustration of the Parameters Involved in Constructing the Final Feature Vector in the Seq-to-Seq Model*

**Algorithm 1.** Training for LETM-AE as a seq2seq learning

Input: N framework, Epoch, Configuration set to LSTM-AE.
Initialization: Parameter set to $\Theta$, $epoch \leftarrow 1$.
Notice: The algorithm is for m=1, code_size=1
1: Extract $F$ feature for each window with HUF-net
2: Making q sequence of for each $F$ as a $X$
// Forward propagation of LSTM-AE
3: for epoch=1, 2, …, Epoch do:
4: for n=1,2, 3, …, N do:
// Encoder
5: for $i_e = 1,2 \dots, q$:

$$F_{i_e} = \sigma(W_{f_{i_e}} * concat\left(h_{f_{i_{e-1}}}, X_{i_e}\right) + b_{f_{i_e}})$$
$$I_{i_e} = \sigma(W_{I_{i_e}} * concat\left(h_{I_{i_{e-1}}}, X_{i_e}\right) + b_{I_{i_e}})$$
$$C_{i_e} = \tanh\left(W_{c_{i_e}} * concat\left(h_{C_{i_{e-1}}}, X_{i_e}\right) + b_{c_{i_e}}\right)$$
$$O_{i_e} = \sigma\left(W_{o_{i_e}} * concat\left(h_{o_{i_{e-1}}}, X_{i_e}\right) + b_{o_{ie}}\right)$$
$$CS_{i_e} = (CS_{i_{e-1}} \times F_{i_e}) + \left(I_{i_e} \times C_{i_e}\right)$$
$$h_{i_e} = O_{i_e} \times \tanh\left[(CS_{i_{e-1}} \times F_{i_e}) + \left(I_{i_e} \times C_{i_e}\right)\right]$$

// Decoder
6: $Z = h_{ie=q} * 0$
7: $x = \left[h_q, Z, Z, \dots, Z\right]_{h_q \times q}$
7: for $i_d = 1,2, \dots, q$:

$$F_{i_d} = \sigma(W_{f_{i_d}} * concat\left(h_{f_{i_{d-1}}}, x_{i_d}\right) + b_{f_{i_d}})$$
$$I_{i_d} = \sigma(W_{I_{i_d}} * concat\left(h_{I_{i_{d-1}}}, x_{i_d}\right) + b_{I_{i_d}})$$
$$C_{i_d} = \tanh\left(W_{c_{i_d}} * concat\left(h_{C_{i_{d-1}}}, x_{i_d}\right) + b_{c_{i_d}}\right)$$
$$O_{i_d} = \sigma\left(W_{o_{i_d}} * concat\left(h_{o_{i_{d-1}}}, x_{i_d}\right) + b_{o_{i_d}}\right)$$
$$CS_{i_d} = (CS_{i_{d-1}} \times F_{i_d}) + \left(I_{i_d} \times C_{i_d}\right)$$
$$h_{i_d} = O_{i_s} \times \tanh\left[(CS_{i_{s-1}} \times F_{i_s}) + \left(I_{i_s} \times C_{i_s}\right)\right]$$

8: $l = \sum_1^q \left(X_i - h_{i_d}\right)^2$
// Adam optimizer [35]
9: Update weights
$W_{f_{i_d}}, W_{I_{i_d}}, W_{c_{i_d}}, W_{o_{i_d}}, b_{f_{i_d}}, b_{I_{i_d}}, b_{c_{i_d}}, b_{o_{i_d}} \leftarrow i_d = q - 1, q, \dots, 1$
$W_{f_{i_e}}, W_{I_{i_e}}, W_{c_{i_e}}, W_{o_{i_e}}, b_{f_{i_e}}, b_{I_{i_e}}, b_{c_{i_e}}, b_{o_{ie}} \leftarrow i_e = q, q-1, \dots, 1$

10: end for
11: end for
Output: Parameter set $\Theta$

### 2.4. Classification

After extracting static features using the HUF method, we classify them using a Dense Layer Network (DLN) classifier to estimate their discriminative capability prior to any temporal enhancement. In the next step, after forming sequences from the extracted features, each framework is classified using a standard LSTM network followed by a DLN classifier. This network, which is structurally designed to resemble the encoder part of the LSTM-AE, will hereafter be referred to as

STM-CF. LSTM-CF allows us to assess the ability of a simple model to improve the representation of a framework using its preceding time steps. Finally, the features extracted by the HUF model are further refined using an LSTM-based autoencoder (LSTM-AE). The resulting code produced by this model is then classified using a DLN classifier. This setup enables a comparison of the framework's separability in three configurations: (1) without any temporal enhancement, (2) enhanced by a simple LSTM, and (3) enhanced via the LSTM-AE. The classification results for each case are reported in the following sections.

## 3. Experiments

It is important to note that although the memory-based autoencoder is trained in an unsupervised manner (i.e., without labels), during evaluation, only the label corresponding to the final vector in each sequence is used for classification. Previous vectors are solely used to enrich the representation of the target window.

### 3.1. Datasets

To evaluate the proposed method, we selected two datasets with relatively high complexity in terms of the number of activity classes and sensor configurations. The first dataset is DaLiAc, which contains data from 19 participants performing 14 activity tasks characterized by high inter-class similarity. The data were collected using four sets of accelerometers and gyroscopes [3]. The second dataset used in this study is PAMAP2, which, despite containing fewer total samples than other datasets, provides rich diversity. In this dataset, nine individuals performed 11 activity tasks recorded via three IMU sensors. Each sensor includes an accelerometer, gyroscope, and magnetometer; however, due to the limited usage of magnetometers in related literature, we have excluded them from our analysis as well; moreover, our experiments confirm the magnetometer does not significantly impact classification performance but does increase computational cost. Table 1 provides detailed information on each dataset, including activity types, sampling rates, and sensor placement locations. Additionally, shows the distribution of samples per class. As illustrated, walking is the most frequent activity in both datasets, with PAMAP2 showing a nearly equal number of samples for ironing.

*Table 2 Detailed specifications of the datasets used in this study*

| Dataset | Participants | Activity | Sensors | Placed sensors | Sample frequency | Window Size | Window overlap |
|---|---|---|---|---|---|---|---|
| DaLiAc | 19 | Sitting, Lying, Standing, Washing dishes, Vacuuming, Sweeping, Walking, Ascending stairs, Descending stairs, Bicycling on ergometer (50 and 100 W), Rope jumping | Acc , Gyro | Right hip Chest Right Wrist Left ankle | 200 Hz | 200, 500 | 50% |
| PAMA P2 | 9 | Lying, Sitting, Standing, Walking, Running, Cycling, Nordic walking, Ironing, Vacuum Cleaning, Rope jumping, Ascending and Descending stairs | Acc, Gyro, Mag, HR | Wrist, Chest, Ankle | 100 Hz (Up sample to 200Hz) | 200, 500, 1000 | 50% |

## 3.2. Segmentation

The strategy used for segmenting continuously sampled time-series signals into windows is a crucial factor that significantly influences the training and overall performance of learning models. In the majority of HAR datasets, data collection is performed continuously, without interruption, even as subjects transition from one activity to another. Consequently, some portions of the recorded signal may correspond to intermediate or ambiguous states in which no specific activity is clearly defined. In this study, without altering the temporal order of the dataset, we isolate only those parts of raw data that are associated with labeled activities. Over each of these parts, we apply a sliding window of two to three seconds, extracting N samples per window. Each window is then labeled based on the activity tag of its last sample. It is important to note that when an activity transition occurs, a window may occasionally contain portions of two different activities.

In contrast to prior work, which typically applies intra-class windowing—where each window contains only one type of activity and labeling is done accordingly—we adopt an inter-class windowing strategy. This approach aligns more closely with real-world scenarios, in which activities frequently overlap or change gradually, making purely intra-class windows unrealistic. Our experimental results show that including such transitions may lead to approximately a seven percent reduction in HAR during classification.

By applying this segmentation approach, we extract $T$ windows from each subject's data, resulting in a four-dimensional tensor with the shape $[T, S, C, N]$, where $S$ is the number of sensors sets, $C$ is the number of channels per sensor and $N$ is the number of samples in each window.

It is also important to note that subject-based hold-out is applied for training and testing splits (70% for training and 30% for test). In other words, data from the same individual is not shared across both sets.

## 3.3. Implementation the models for DaLiAc datasets

### *3.3.1. Data Representation*

To assess the impact of LSTM-AE parameter variations on the model's ability to accurately recognize different human activity classes, we first evaluate the proposed model on the DaLiAc dataset using various parameter settings. Also, we visualize several activity classes from the DaLiAc dataset for subject number 3 in Figure 8, to demonstrate the relationship between the model and the raw motion data.

In Figure 8, we plot the accelerometer and gyroscope signals from the subject's wrist sensor over the sample range 500 to 1500, corresponding to six different activity classes. Within this five-second window, walking occurs repeatedly, while activities like sweeping and cycling on an ergometer at 100 watts appear only twice, and lying down is observed once. These observations highlight the differing temporal dynamics across activity types, which is particularly important when using short window durations in HAR tasks. For example, incorporating prior motion context for the sweeping activity greatly aids in distinguishing it from sitting and lying down, as all three activities exhibit minimal acceleration changes during their first second.

### 3.3.2. Parameter Settings

#### 3.3.2.1. HUF Method

The DR-SAE, LFF-AE, and GFF-AE networks parameters are configured as illustrated in Figure 2 and detailed in Table 1. The kernel size for the convolutional layers in all three networks is set to 1×3. Depending on the sampling rate of the dataset and the number of samples in each input framework ($N$), the kernel sizes of the pooling layers are adjusted accordingly to ensure accurate reconstruction at the output of the decoder. After extracting the static features by the HUF model, the features are flattened and passed through a four-layer dense neural network for classification.

#### 3.3.2.2. Dynamic Networks

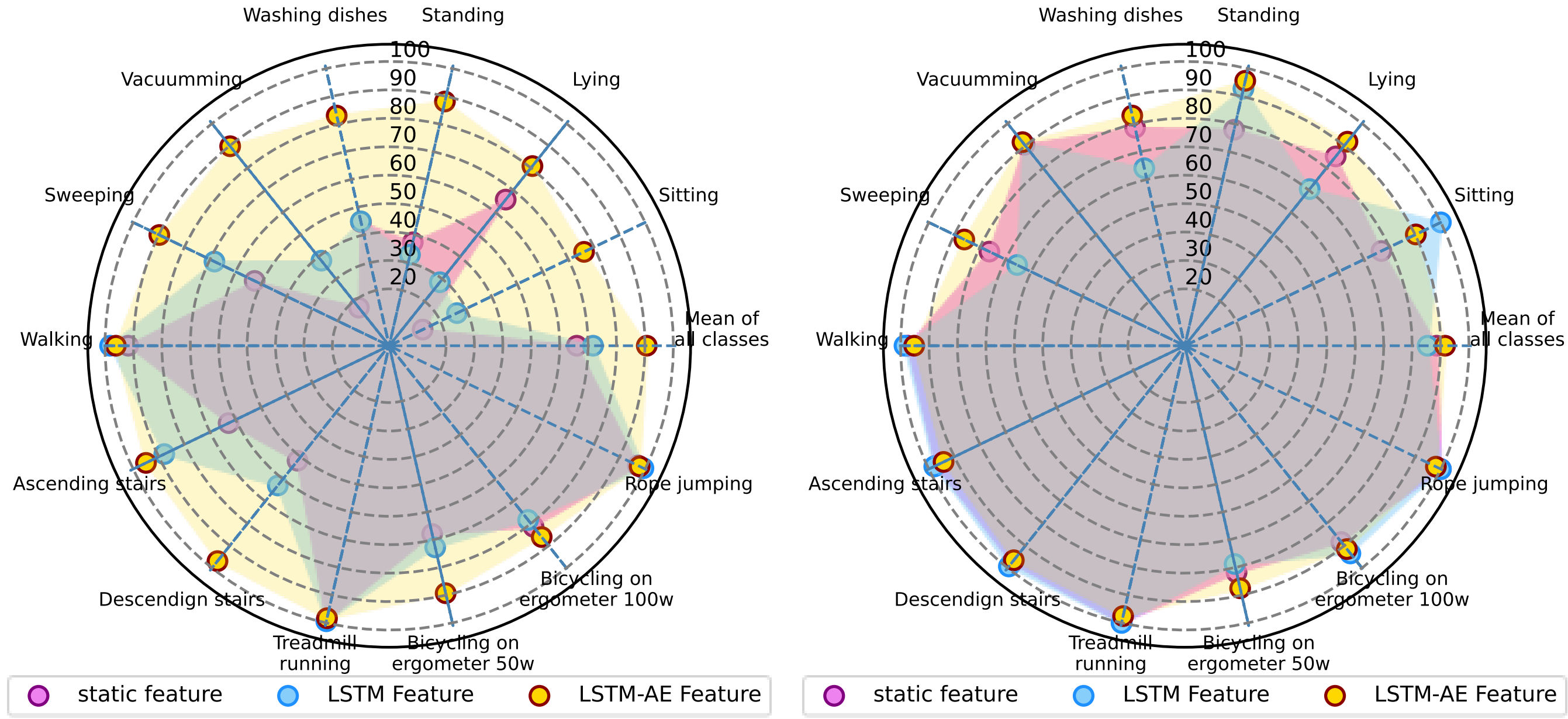

*Figure 7 Classification results of spatial features extracted by the HUF model and enhanced feature using the proposed LSTM-AE model for two framework lengths: N=512 (right) and N=200 (left) on the DaLiAc*

In the dynamic feature extraction stage, the goal is to enhance the representation of the current window feature on information from previous windows features. Firstly, a sequence of input motion windows is formed as illustrated in Figure 3. Secondly, the obtained sequence is processed and classified using a LSTM-CF network. As explained in 2.4, the LSTM network consists of six cells, and the output of the final cell is followed by a four-layer dense network. Finally, according to the results in Section III.D, the optimal configuration for the LSTM-AE model is achieved with a single-layer encoder, $q = 6$, $m = 1$, and a *code size* of *1000*. Under these conditions, the code layer provides the most discriminative feature representation for classifying human activity classes using a four-layer dense classifier. We present the classification results of features extracted by the **HUF**, **LSTM-CF**, and **LSTM-AE** models in Figures 9, using two different framework lengths from the DaLiAc dataset. In both figures, in addition to reporting the average classification accuracy, we also show the per-class performance. As demonstrated, the feature enhancement achieved by the LSTM-AE model leads to the highest classification accuracy.

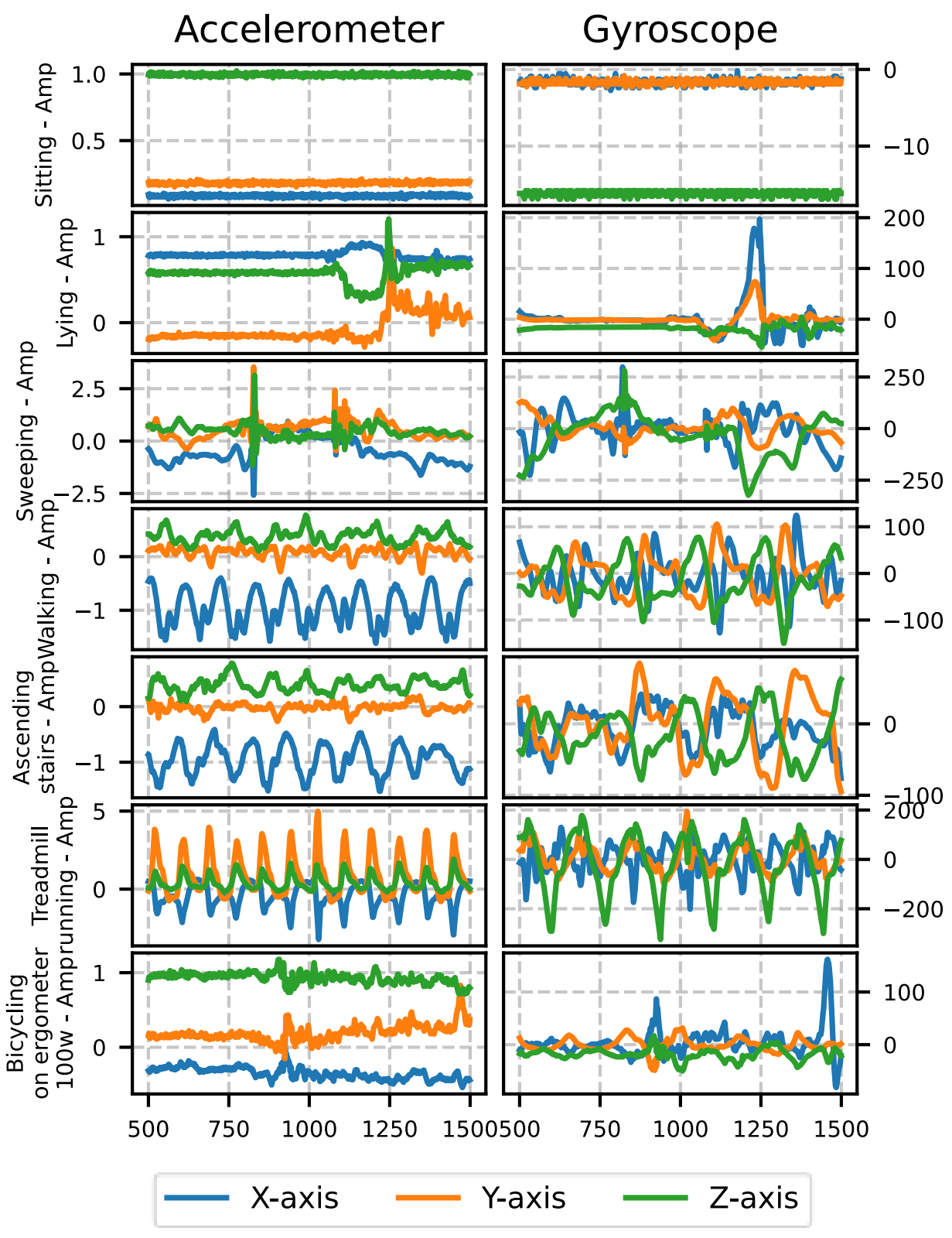

*Figure 8 Sample segments of multiple wrist sensor activities from the DaLiAc dataset for subject number 3*

Specifically, for $N = 200$ (i.e., a 1-second window), the LSTM-AE method improves classification separability by approximately 20% compared to the other two methods. Under this setting, features from the HUF model (pink color) are insufficient to distinguish between certain classes such as standing and vacuuming, and the average classification accuracy across all classes is 63%. The LSTM-CF model (purple color) increases accuracy to 73%, but still falls short of a robust result. However, after enhancing the HUF features using the proposed LSTM-AE model, the accuracy improves significantly to 96.6%. This not only enables the accurate recognition of the aforementioned classes but also pushes some classes toward near-perfect classification performance.

When using $N = 512$ (i.e., a 2.56-second window), the baseline features extracted by the HUF model are already improved, yielding a classification accuracy of over 86%. In this case, all activity classes are learned to an acceptable level, with no class left unrecognized. Although the static features alone achieve satisfactory results, applying LSTM-CF raises accuracy to 87.3%, and applying LSTM-AE further boosts it to 91.6%. A particularly notable outcome of this analysis is the substantial improvement observed for the $N = 200$ samples in a window. Despite the limited discriminative power of the static features in this short window, the **LSTM-AE** enhancement elevates their performance to surpass even the longer window ($N = 512$). This highlights the effectiveness of the proposed model in enhancing low-resolution features through short-term temporal context.

## 3.4. Effect of Hyperparameters on STS Feature Development in the DaLiAc Dataset Using LSTM-AE

### *3.4.1. Varying sequence length and Code size*

In this section, we examine the effect of the hyperparameters sequence length ($q$), code size, and encoder-decoder cell connection size ($m$) in the LSTM-AE model. To isolate their individual impact, we vary each parameter while keeping the other two fixed. For all experiments in this section, the encoder and decoder are configured with a single LSTM layer. After identifying the optimal values for $q$, code size, and $m$, we will later evaluate the impact of increasing the number of layers.

It is well known that increasing $q$, extends the length of the input sequence, resulting in more LSTM cells and, consequently, a larger number of trainable parameters. Similarly, increasing the code size directly enlarges the number of weights within each LSTM cell. To evaluate the impact of these parameter changes on model performance, we tested $q = 3,6,10,14$ that for each of which the code sizes is $500, 1000, and\ 3000$ and the results are presented in Figure 10. As shown, the highest classification accuracy is achieved at $q = 6$, indicating that this sequence length yields the most effective temporal representation in the LSTM-AE model. Furthermore, for $q = 6$, the best performance is obtained when the code size is 1000, outperforming both smaller and larger configurations. The figure also suggests that increasing the number of LSTM cells ($q > 6$) results in higher accuracy when smaller code sizes are used. This may be attributed to the size of the training dataset, as larger values of $q$ increase the number of model parameters. Thus, using a smaller code size helps control model complexity and facilitates better training. Notably, increasing the number of neurons in the encoder output (code size) has a greater impact on the total parameter count than increasing the sequence length alone.

It is also worth noting that for $q = 3$, the total temporal span of each input sequence is less than 1.5 seconds. Based on the motion examples illustrated in Figure 8, such a short context may be adequate for periodic actions like walking or running, where a complete cycle can be captured. However, for more extended activities such as sweeping or lying down, longer temporal dependencies are necessary. This reinforces the importance of selecting an appropriate $q$ value to capture the motion dynamics of longer-duration activities during training.

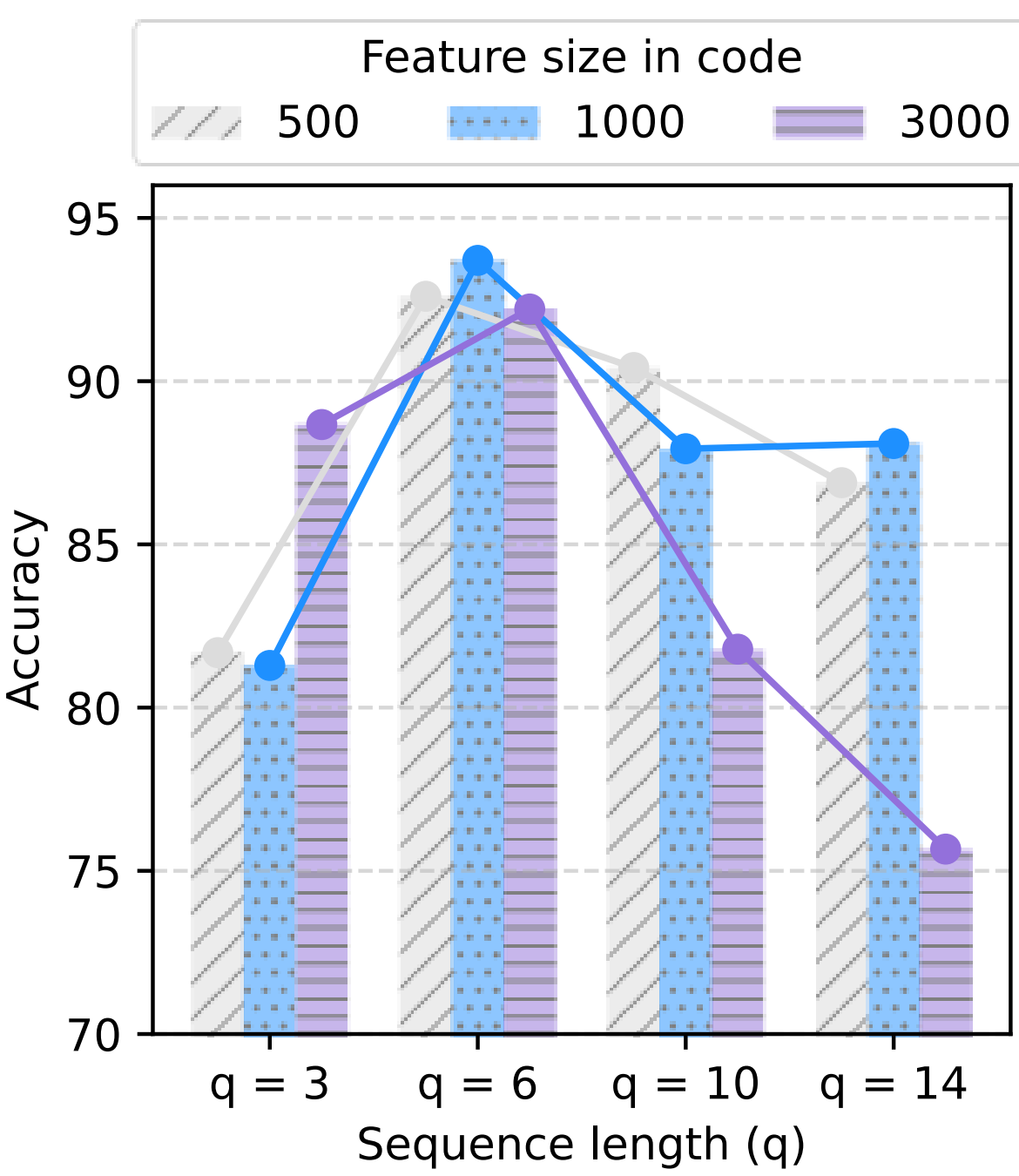

*Figure 9 The impact of varying the number of cells in the encoder–decoder sequence (q) and the code size on the classification performance of the LSTM-AE model.*

### 3.5. Effect of Changing mmm Length

Another parameter examined in this study is the number of encoder–decoder connection cells, denoted by $m$, which was tested at values of $m = 1, 2, \ and\ 3$, while keeping $q = 6$ and the code size fixed at 1000. During this evaluation, various configurations of the updated feature representations were analyzed, and the corresponding classification results are presented in Table 2. For instance, when $m = 2$, we evaluated three different feature configurations for classification: (1) using only the final cell, (2) using the cell preceding the final one, and (3) using a concatenation of the last two cells.

As shown in Table 2, increasing the value of $m$ leads to a decline in feature separability, with a notable drop in classification performance at $m = 3$. For $m = 2$, three different code-layer selection strategies were considered, yet in all cases, the classification accuracy remained below 89%. At m=3, the highest observed accuracy was 78.53%, achieved when only the features from the final LSTM cell were used.

A noteworthy observation is using all features from the encoder–decoder connecting cells (i.e., $m > 1$) does not improve the model's performance, despite the expectation that aggregated features would retain more temporal information than a single cell. However, the increased dimensionality of these combined features significantly raises the number of parameters in the classification network, which—given the limited training data—may lead to overfitting or suboptimal learning.

The results indicate that a larger number of $m$ tends to diminish the quality of feature representations in the proposed LSTM-AE framework. This observation is consistent with established principles of autoencoder design, in which limiting the bottleneck capacity promotes stronger compression and yields features that are more discriminative for downstream classification tasks.

*Table3 Analysis of the effect of varying the number of encoder–decoder connection cells on the features extracted by the LSTM-AE model. Here, **q** indicates the index of the LSTM cell whose output was used for feature extraction.*

| *m* | *Output features for Classification* | *Number of features* | ***Accuracy of classification features in test data*** |
|---|---|---|---|
| *1* | *cell = 6* | *1000* | ***93.7*** |
| *2* | *cell = 6* | *1000* | ***88.04*** |
| | *cell = 5* | *1000* | ***88.38*** |
| | *cell = 5,6* | *2000* | ***88.17*** |
| *3* | *cell = 6* | *1000* | ***78.53*** |
| | *cell = 5* | *1000* | ***78.34*** |

| | cell = 4 | 1000 | **76.79** |
|---|---|---|---|
| | cell = 5,6 | 2000 | **77.78** |
| | cell = 4,5,6 | 3000 | **77.58** |

### 3.6. Changing the Number of Layers

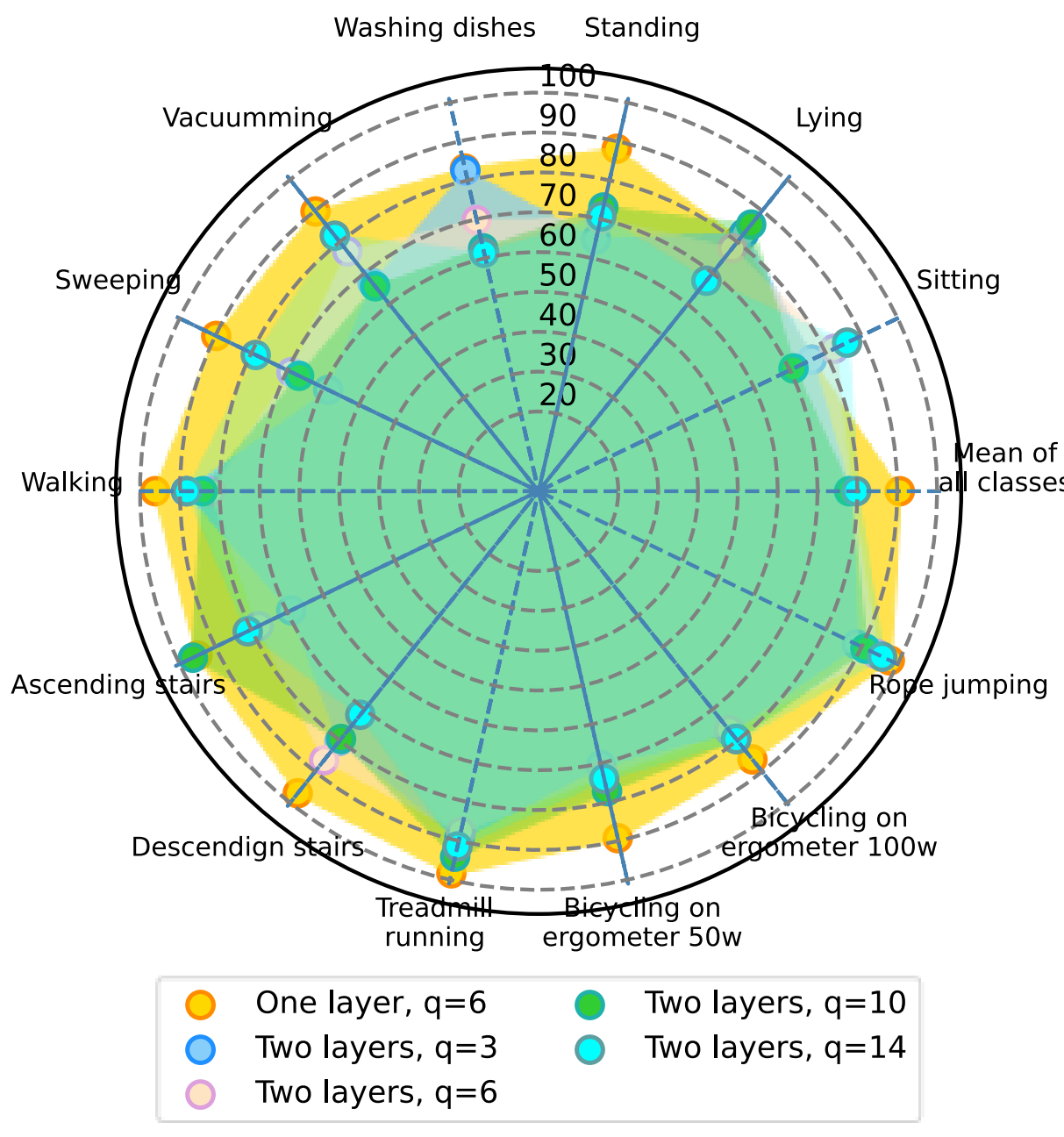

*Figure 10 Comparison of the classification performance of enhanced features extracted using the LSTM-AE model with single-layer and two-layer encoder configurations, across different numbers of LSTM cells (q=3 to 14).*

The results discussed in the previous sections were based on a single-layer encoder in the LSTM-AE sequence. In this section, we increase the number of encoder layers to two, while keeping the parameters fixed at $code\ size = 1000$ and $m = 1$. We then evaluate feature enhancement for different values of $q = 3, 6, 10, 14$ and analyze the effect of increasing encoder depth alongside the number of LSTM cells. Although adding more encoder layers increases the number of parameters in the recurrent model, but it is not doubled. This is because the input to the first layer is 2240 neurons, while the input to the second layer corresponds to the output of the first layer (i.e., 1000 neurons), which proportionally reduces the number of hidden units required in the second LSTM layer.

As shown in Figure 11, increasing the encoder depth generally leads to a decline in overall classification performance. However, for certain activity classes such as sitting and washing dishes, which exhibit longer or more complex dynamics, the two-layer encoder improves performance. Furthermore, under the two-layer setting, increasing the number of LSTM cells does not significantly affect the average classification accuracy across all classes. While the ability to distinguish some classes improves, performance on others may decline, resulting in same classification accuracy.

## 4. Results

### 4.1. DaLiAc Dataset

To evaluate the effectiveness of feature enhancement using the LSTM-AE network, we compare its results with two baseline methods described in section 2.4, as well as with existing approaches from prior literature. It is important to note that, unlike most previous studies, this work adopts an inter-class segmentation strategy, where the signal windows may contain transitions between different activities. As discussed in section 2.4 and showed in Table-4, this realistic segmentation method leads to approximately a 7% drop in classification accuracy for method such as HUF. However, it reflects a more practical and application-oriented setting.

**Table 4** presents a comparison of classification results obtained using the proposed **LSTM-AE** model and other HAR methods applied to the DaLiAc dataset. Since the field of HAR lacks extensive research on unsupervised or semi-supervised models, most reported methods are fully supervised. However, these supervised approaches suffer from key limitations, including a strong dependence on labeled data and reduced robustness when class distributions are imbalanced; as noted in the Introduction and discussed in [36 ,31].

*Table4 Comparison of classification accuracy for 13 activity classes from the DaLiAc dataset using various feature extraction methods, including supervised approaches and our unsupervised method. The proposed approach evaluates feature extraction both without considering motion dynamics and with dynamic feature enhancement using the STS-based model and the proposed LSTM-AE network.*

| Method | Framework time | Acc |
|---|---|---|
| Deep3DCNN [37] | 5s | **94.9** |
| Iss2Immage + 2D-CNN [38] | 5s | **92.6** |
| Signal visualization + Deep CNN-LSTM [37, 39] | - | **90.2** |
| Probabilistic compact feature + KNN [37, 40] | 5.12s | **90.3** |
| LSTM [37, 40] | 5.12s | **88.9** |
| Hierarchical Multiple Classifier [3] | 5s | **89.6** |
| | | |
| **HUF method** (inter-HA segmentation) | 2.56s | **87.6** |
| **HUF method** (intra-HA segmentation) | 2.56s | **93.8** |
| **LSTM-CF** (inter-HA segmentation) | **1s** | **89.3** |
| **LSTM-AE (STS method)** Proposed method (inter-HA segmentation) | **1s** | **96.6** |

### 4.2. PAMAP2 Dataset

Based on the findings in Section 3.4, we applied the optimal parameters obtained from the DaLiAc dataset to the PAMAP2 dataset for feature enhancement using the proposed STS-based LSTM-AE model. The results of this implementation, along with those of other methods, are presented in Table 3. It is important to note that many of the comparison methods listed in the table use larger temporal windows than the one adopted in our proposed model. In some cases, the window length used in prior studies is impractical for real-time activity recognition. Moreover, unlike previous work, this study applies inter-class segmentation across the entire raw dataset, introducing a more realistic and challenging evaluation setting. Despite these constraints, the LSTM-AE model effectively enhances the static features extracted by the HUF model, achieving results comparable to or better than those of prior methods. This demonstrates the robustness of the proposed approach, even under tighter temporal windows and more challenging data conditions.

*Table5 Comparison of classification accuracy for 12 activity classes in the PAMAP dataset using enhanced features from the proposed LSTM-AE method versus motion features extracted by other existing approaches.*

| Method | Framework time | **Acc** |
|---|---|---|
| Multi-Branch CNN-GRU with attention [41] | 10s | **98.6** |
| LSTM [42] | 10s | **97.64** |
| Transformer-LSTM [42] | 10s | **96.3** |
| Semi-Recurrent CNN-LSTM attention model [36] | 5s | **83.42** |
| | | |
| **LSTM-AE (STS method)** Proposed method (inter-HA segmentation) | **1s** | **98.4** |

## 5. Discussion

Challenges such as data scarcity, diverse activity patterns, and the integration of multiple sensors, particularly across different activity classes, have limited the widespread adoption of unsupervised models in HAR. Furthermore, modeling the temporal dynamics of motion is a core requirement in HAR; however, training such models typically requires large labeled datasets. In contrast, unsupervised methods offer the potential to accurately analyze human motion across diverse populations, including variations in gender, weight, and age.

In this study, we addressed these issues by proposing an unsupervised model that enhances static motion features, extracted from a short-term activity window using information from preceding windows. First, we employed a hierarchical unsupervised method [31] to extract spatial features from a single window. Then, we introduced a memory-based autoencoder (LSTM-AE) built on recurrent networks to enhance these features using temporal sequences, following an STS learning approach. Importantly, the proposed model operates on shorter temporal windows than previous studies and demonstrates improved performance on both the DaLiAc and PAMAP2 datasets. In other words, despite imposing more challenging conditions than earlier research, such as using smaller window sizes and inter-class segmentation that may include multiple activity types within a single frame, the model consistently achieves higher accuracy.

We demonstrated experimentally that inter-class segmentation alone can reduce classification accuracy by around 7% in static unsupervised feature extraction models. Despite this, our LSTM-AE model was able to learn and accurately classify all 13 activity classes in the DaLiAc dataset using just a 1-second window. In contrast, under static conditions, the model struggled to distinguish between similar classes such as sitting, lying down, and vacuuming.

Additionally, while increasing the window size generally improves both static and dynamic feature quality, we observed that overly long sequences lead to decreased final performance in the LSTM-AE model. This drop is primarily due to the mismatch between extended sequence length and natural human movement cycles, as well as the rapid increase in model complexity (i.e., number of parameters), which affects performance.

For the PAMAP2 dataset, although our model did not surpass all existing supervised methods in terms of raw accuracy, it was evaluated under much stricter and more realistic conditions. Specifically, the temporal window used was only 10% the length of those in some comparative studies, and inter-class segmentation was applied throughout. The robustness of our model under these constraints highlights its practical relevance.

While the LSTM-AE effectively addresses many of HAR's key challenges through its unsupervised architecture and outperforms others under more realistic settings, it does come with a trade-off: longer training times due to the STS architecture. This may require more powerful hardware for efficient deployment